\documentclass{article}
\usepackage[preprint]{styles} 
\usepackage{amsmath}
\usepackage{graphicx}
\usepackage{amsfonts}
\usepackage{booktabs}
\usepackage{subcaption}
\usepackage{float}
\usepackage{multirow}
\usepackage{placeins}
\setcitestyle{numbers,square}

\title{Biological Pathway Informed Models with Graph Attention Networks (GATs)}
\workshoptitle{AI for Science: The Reach and Limits of AI for Scientific Discovery}

\date{August 2025}

\author{
  Gavin Wong\\
  Yale University\\
  \texttt{gavin.wong@yale.edu}
  \And
  Ping Shu Ho\\
  NVIDIA AI Tech Center\\
  \texttt{cliffh@nvidia.com}
  \And
  Ivan Au Yeung\\
  NVIDIA AI Tech Center\\
  \texttt{iauyeung@nvidia.com}
  \And
  Ka Chun Cheung\\
  NVIDIA AI Tech Center\\
  \texttt{chcheung@nvidia.com}
  \And
  Simon See\\
  NVIDIA AI Tech Center\\
  \texttt{ssee@nvidia.com}
}

\begin{document}

\maketitle

\begin{abstract}

Biological pathways map gene–gene interactions that govern all human processes. Despite their importance, most ML models treat genes as unstructured tokens, discarding known pathway structure. The latest pathway-informed models capture pathway-pathway interactions, but still treat each pathway as a ``bag of genes" via MLPs, discarding its topology and gene-gene interactions. We propose a Graph Attention Network (GAT) framework that models pathways at the gene level. We show that GATs generalize much better than MLPs, achieving an 81\% reduction in MSE when predicting pathway dynamics under unseen treatment conditions. We further validate the correctness of our biological prior by encoding drug mechanisms via edge interventions, boosting model robustness. Finally, we show that our GAT model is able to correctly rediscover all five gene-gene interactions in the canonical TP53-MDM2-MDM4 feedback loop from raw time-series mRNA data, demonstrating potential to generate novel biological hypotheses directly from experimental data. 

\end{abstract}

\section{Introduction}

Biological pathways are the fundamental logic the human body is built on. Our extensive knowledge on pathways can serve as an effective prior to guide models to learn true biological relationships and avoid noise. However, existing ML models often treat genes as unstructured tokens, or at most encode interactions at the pathway level. To improve on this, we make the following contributions:

\begin{enumerate}

\item {We propose a novel method to explicitly encode biological pathways at the gene level using Graph Attention Networks (GATs).}
\item {We show that encoding known biological pathways as a mechanistic prior allows models to learn a more robust, interpretable, and generalizable set of pathway dynamics.}
\item {We suggest the potential of our GAT formulation to discover new biological insights, such as candidate pathways and novel gene-gene interactions.}

\end{enumerate}

\section{Related Work}

The latest approaches to incorporating biological pathway knowledge into ML models include: (1) encoding gene-pathway membership via sparse neural networks \cite{xie2024tracing, cai2024deepathnet}; and (2) encoding pathway-pathway interactions using attention biases or graph priors \cite{liu2024pathformer, dong2023highly, ma2024graphpath}. While these approaches have achieved superior results on many downstream tasks, they remain limited by only encoding interactions at the pathway level, discarding the structured gene-gene interactions that define a pathway. For example, in the canonical p53 pathway, the gene TP53 activates MDM2, while MDM2 in turn inhibits TP53, forming a negative feedback loop that is critical to regulating p53. Treating TP53 and MDM2 as independent tokens aggregated into a single "p53 pathway" node loses this mechanistic detail.

We address this by proposing a Graph Attention Network (GAT) method to  encode the natural graph structure of a biological pathway. We encode genes as nodes, and use multiple adjacency matrices to represent different interaction types (e.g. activatory/inhibitory). Below, we show: 

\begin{enumerate}

\item{Better generalization of our GAT model to unseen treatment conditions versus MLP;}

\item{Correctness of our biological prior via edge interventions reflecting drug mechanisms; and}

\item{Ability to rediscover known biology, suggesting potential to generate novel insights.}

\end{enumerate}

\section{Methods}

\subsection{Data}

We use our framework to model the core feedback loop of the p53 pathway, composed of the genes TP53, MDM2, and MDM4 (Figure 1). TP53 is a tumor suppressor whose activity is tightly regulated by MDM2 and MDM4 via negative feedback. This feedback loop generates oscillatory dynamics, presenting a richer challenge than linear signaling pathways.

\begin{figure}
    \centering
    \includegraphics[width=\linewidth]{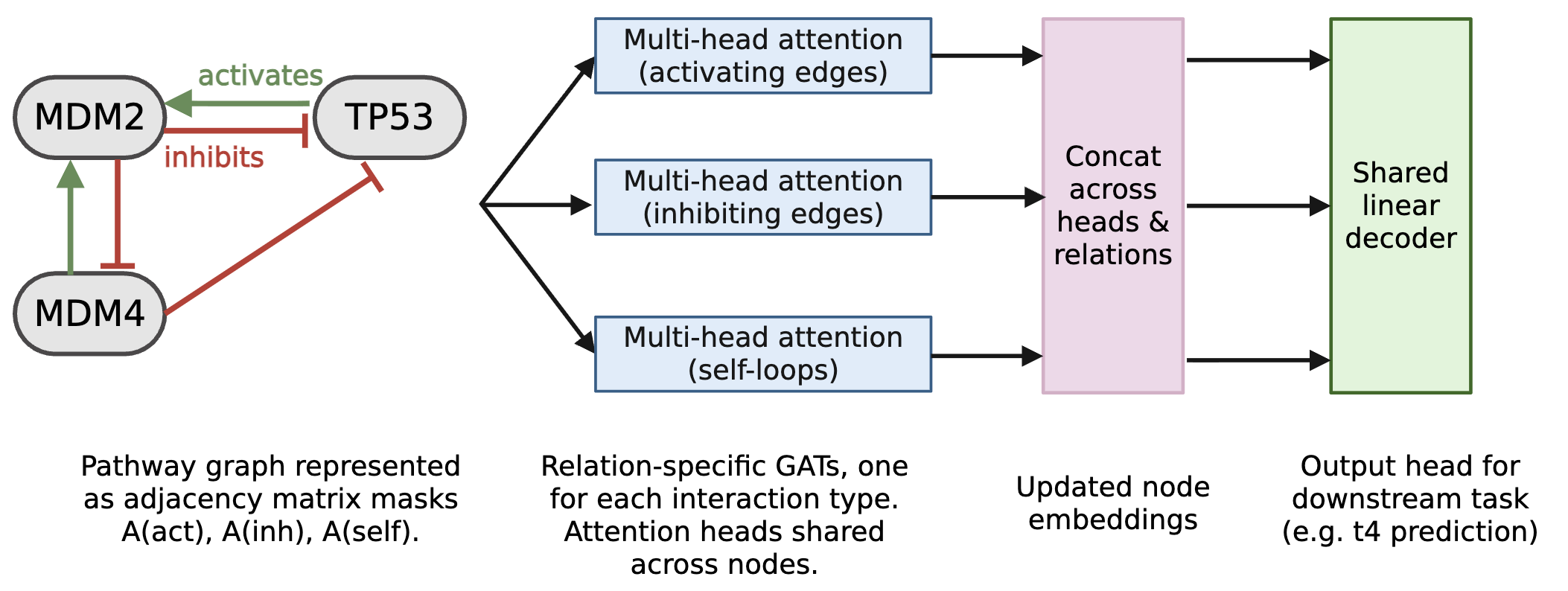}
    \caption{{Model schematic}}
    \label{fig:placeholder}
\end{figure}

\textbf{Dataset.} We use time-series mRNA expression for TP53, MDM2, and MDM4 from Hafner et al. \cite{hafner2017p53} (GEO: GSE100099). Experiments cover three treatment conditions: \textbf{Wild-type (WT)} with no intervention; \textbf{TP53-sh}, a knock-down that dampens TP53 expression; and \textbf{Nutlin}, a drug that blocks the MDM2-TP53 interaction. For each condition there are two independent trajectories; within a trajectory, the three genes are measured at 6-12 time points over 0-24 hours.

\textbf{Problem formulation.} Given measurements at \(t_1\), \(t_2\), \(t_3\) plus metadata (elapsed time between measurements \(\Delta t_{1:3}\) and treatment indicators), predict the expression at \(t_4\) for the three genes. Samples were generated by sliding a \(t_1, t_2,t_3 \rightarrow t_4\) window along each trajectory.

\textbf{Evaluation protocol (LOCO).} To test model robustness, we adopt Leave-One-Condition-Out (LOCO) validation. Each fold trains on two treatment conditions, and is then evaluated on the held-out condition. This assesses the model's ability to learn pathway dynamics that generalize to unseen treatment conditions. Unless stated otherwise, we report mean squared error (MSE) on the standardized target space averaged over 10 random seeds (mean \(\pm\) std) and compare against an MLP baseline that reflects the "bag-of-genes" approach used in the literature. 

\subsection{Model Architecture}

We build a pathway graph with \(N\) nodes (one per gene). Node \(i\) receives the feature vector:

\begin{equation}\label{eq:one}
\mathbf{x}_i = [\; \mathbf{y}_i(t_{1:3}) \; || \; \Delta \mathbf{t}_{1:3} \; || \; \mathbf{u} \;]
\end{equation}

where \(\mathbf{y}_i(t_{1:3}) \in \mathbb{R}^3\) are the first three expression measurements of gene \(i\) in the current window,  \(\Delta \mathbf{t}_{1:3} \in \mathbb{R}^3\) are the corresponding inter-measurement time gaps (identical for all nodes), and \(\mathbf{u} \in \{0, 1 \}^K\) is the treatment indicator vector (also identical for all nodes). 

Next, we encode pathway structure with a set of relation types \(\mathcal{R}\) (here, \(\mathcal{R}\) = \(\{\)\textit{activatory}, \textit{inhibitory},\textit{ self-loops}\(\}\)). For each \(r \in \mathcal{R}\), we define a binary adjacency \(\mathbf{A}^{(r)} \in \{0, 1\}^{N \times N}\), where \(\mathbf{A}^{(r)}_{ij} = 1\) means that, under relation \(r\), gene \(j\) can attend to gene \(i\). For example, the relation "TP53 activates MDM2" is represented as \(\mathbf{A}^{(\text{\textit{activatory}})}_{\text{\textit{MDM2}}, \text{\textit{TP53}}} = 1\). We use raw node features as embeddings: \(\mathbf{h}_i \xleftarrow{} \mathbf{x}_i\).

For each relation \(r \in \mathcal{R}\), we run a GAT block with \(H\) heads. Each head \(h\) learns its own projection matrix \(\mathbf{W} \in \mathbb{R}^{F \times D}\) and node-level attention matrix \(\mathbf{a} \in \mathbb{R}^{2D}\). We compute attention scores as:

\begin{equation}\label{eq:two}
\alpha_{ij} = \text{\textit{softmax}}_{j \in \mathcal{N}^{(r)}_i}(\text{\textit{LeakyReLU}}(\mathbf{a}^T[\mathbf{W}{\mathbf{h}_i} \; || \; \mathbf{W}{\mathbf{h}_j}]))
\end{equation}

where the softmax function is applied over all permitted neighbors of node \(i\) under relation \(r\), \(\mathcal{N}^{(r)}_i = \{j:A^{(r)}_{ij} = 1\}\). The output feature of node \(i\) is thus:

\begin{equation}\label{eq:four}
\mathbf{h}'_i = \sum_{j \in \mathcal{N}^{(r)}_i} \alpha_{ij}\mathbf{Wh}_j
\end{equation}

We concatenate outputs across the heads of each relation,  then concatenate across relations, generating our aggregated node embedding \(\mathbf{z}_i \in \mathbb{R}^{|\mathcal{R}|HD}\). A linear readout \(\mathbf{W}_{dec} \in \mathbb{R}^{|\mathcal{R}|HD \times 1}\) then maps each node's aggregated embedding to its predicted \(t_4\) expression. Dropout is applied on features and attention weights, and all \(\mathbf{W}^{(r,h)}\) and \(\mathbf{a}^{(r,h)}\) are randomly initialized using the Xavier method. 

\section{Results}

\subsection{GAT model generalizes to unseen treatment conditions much better}
\begin{table}[H]
\caption{\textbf{LOCO performance, GAT vs MLP} (10 seeds)}
\centering
\small
\begin{tabular}{lcc}
\toprule
\textbf{Fold (hold-out)} & \textbf{MLP MSE (mean$\pm$std)} & \textbf{GAT MSE (mean$\pm$std)} \\
\midrule
1 (WT)      & 1.07 $\pm$ 0.02 & 0.44 $\pm$ 0.13 \\
2 (TP53-SH) & 4.92 $\pm$ 0.13 & 0.31 $\pm$ 0.18 \\
3 (Nutlin)  & 50.4 $\pm$ 1.6  & 10.0 $\pm$ 4.1 \\
\midrule
\textbf{Overall mean} & \textbf{18.8 $\pm$ 0.6} & \textbf{3.57 $\pm$ 1.46}\\
\bottomrule
\end{tabular}
\label{tab:loco}
\end{table}

Under LOCO cross-validation, the GAT model achieved 81\% lower overall MSE than the MLP baseline (Table 1), showing that the pathway prior enables much stronger generalization to unseen treatment conditions. The Nutlin case illustrates this best: Nutlin blocks MDM2 from degrading the TP53 protein; elevated TP53 then drives continuous MDM2 transcription, producing a monotonic rise up to 15 standard deviations above normal expression seen in training data (Fig. 2, blue). Without a pathway prior, the MLP fails entirely at predicting this behavior (Fig. 2a). In contrast, the GAT recognizes the feedback loop has been disrupted, and captures the rising trajectory (Fig. 2b-c). 

\begin{figure}[t]
    \centering
    \begin{subfigure}[t]{0.32\textwidth}
        \centering
        \includegraphics[width=\linewidth]{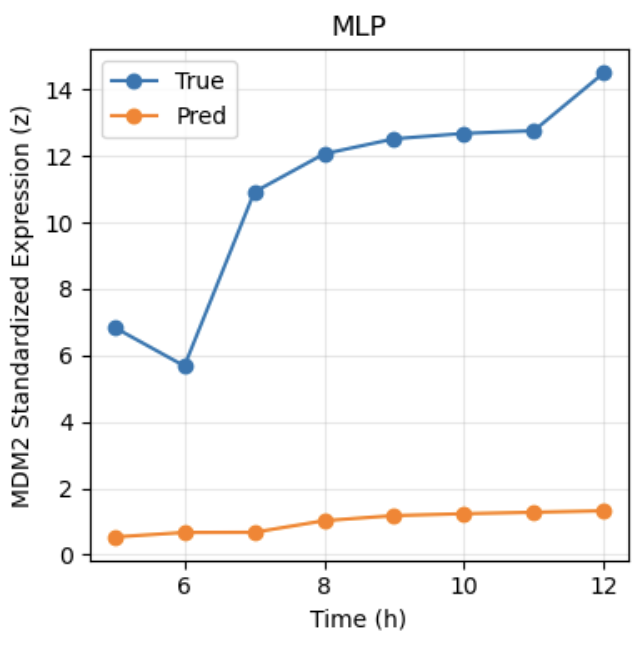}
        \caption{MLP}
        \label{fig:2a}
    \end{subfigure}
    \hfill
    \begin{subfigure}[t]{0.32\textwidth}
        \centering
        \includegraphics[width=\linewidth]{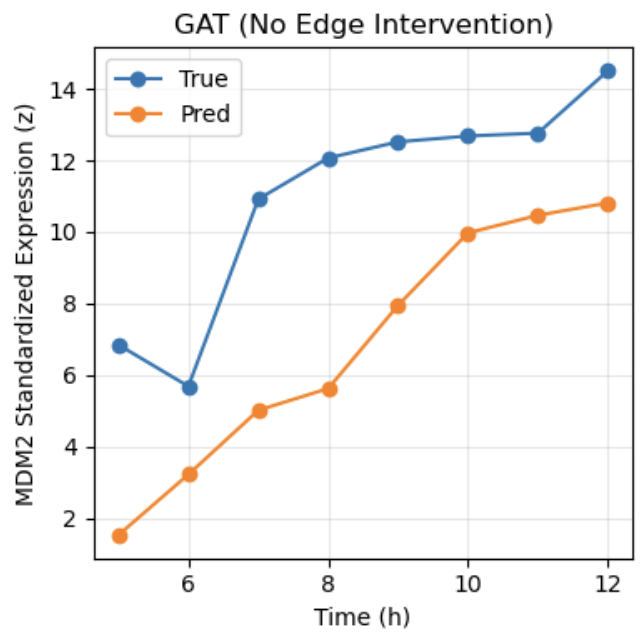}
        \caption{GAT, unmodified pathway}
        \label{fig:2b}
    \end{subfigure}
    \hfill
    \begin{subfigure}[t]{0.32\textwidth}
        \centering
        \includegraphics[width=\linewidth]{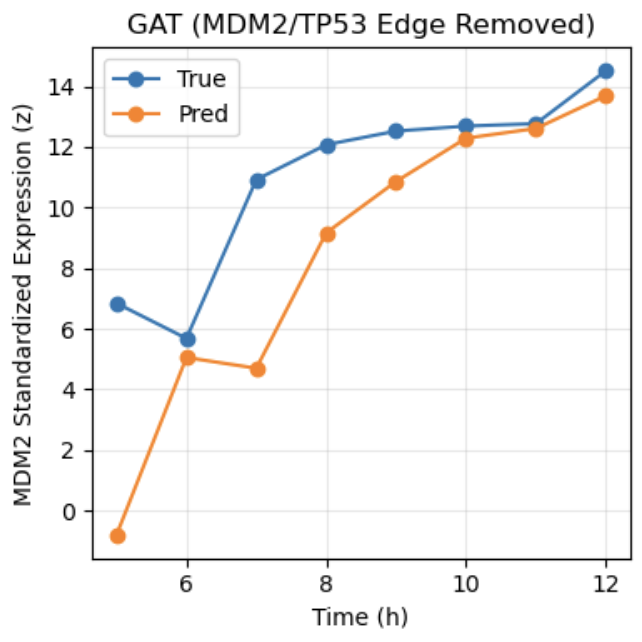}
        \caption{GAT, drug target edge removed}
        \label{fig:2c}
    \end{subfigure}
    
    \caption{True vs Predicted MDM2 expression under Nutlin}
    \label{fig:fig3}
\end{figure}

\subsection{Encoding the drug mechanism via edge intervention improves performance}
\begin{table}[H]
\caption{\textbf{LOCO performance, Drug Mechanism Encoding} (10 seeds)}
\centering
\small
\begin{tabular}{lcc}
\toprule
\multirow{2}{*}{\textbf{Fold (hold-out)}} & \multicolumn{2}{c}{\textbf{MSE (mean $\pm$ std)}} \\
 & \textbf{Unmodified Pathway} & \textbf{Edge Intervention} \\
\midrule
1 (WT)      & 0.52 $\pm$ 0.13 & 0.44 $\pm$ 0.13 \\
2 (TP53-SH) & 0.37 $\pm$ 0.17 & 0.31 $\pm$ 0.18 \\
3 (Nutlin)  & 11.0 $\pm$ 6.8  & 10.0 $\pm$ 4.1 \\
\midrule
\textbf{Overall mean} & \textbf{3.97 $\pm$ 2.24} & \textbf{3.57 $\pm$ 1.46} \\
\bottomrule
\end{tabular}
\label{tab:loco}
\end{table}

The correctness of the biological prior is further shown via an edge intervention. In Fig. 2b, the prior is unchanged; in Fig. 2c, we explicitly model Nutlin's  mechanism of blocking the MDM2-TP53 interaction by setting \(\textbf{A}_{\text{\textit{TP53, MDM2}}}^{\text{\textit{inhibitory}}}=0\). We report a further 11\% improvement in prediction accuracy, demonstrating the GAT model's ability to explicitly incorporate known drug mechanisms. 

\subsection{Given no prior, GAT model rediscovers p53 pathway}

Finally, we trained a GAT with a single, fully-connected adjacency matrix, using the \(tanh()\) activation function to allow negative attention scores. From just raw time-series mRNA data, the model correctly recovered the signs of all 5 gene-gene interactions (Fig. 3). Their relative magnitudes also match known biology (e.g. TP53 being the main activator of MDM2; MDM2 being the main inhibitor of TP53). This suggests the GAT model can also be used to generate novel biological hypotheses (e.g. new pathways/interactions) when trained on raw experimental data without a prior. 

\begin{figure}[H]
    \centering
    \includegraphics[width=0.9\linewidth]{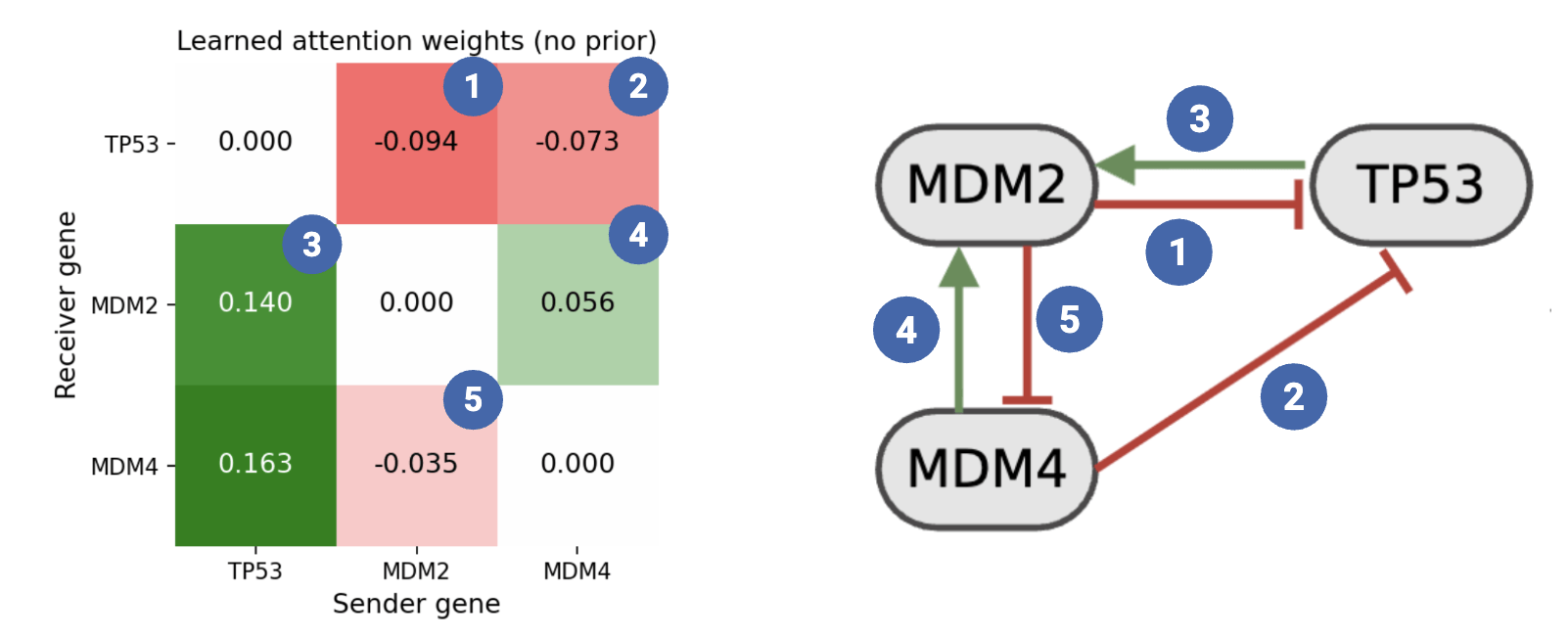}
    \caption{{Learned attention weights (no prior) vs Ground truth pathway graph}}
    \label{fig:placeholder}
\end{figure}

\subsection{Limitations} 

Our study focuses on a single pathway with few genes, and relies on a relatively limited dataset. Nonetheless, our results strongly suggest that encoding pathways at the gene level with GATs yields substantial improvements over MLP baselines, warranting further exploration. 

\FloatBarrier

\section{Conclusion}

We introduced a Graph Attention Network (GAT) framework that encodes biological pathways at the gene level, serving as a mechanistic prior. Our approach generalizes substantially better than existing MLP methods, and offers interpretability to encode known drug mechanisms and  generate novel biological insights. Extensions include encoding temporal dynamics (e.g. via a GRU), and building a multi-pathway, hierarchical framework that uses one GAT layer to map gene data to pathways; and a further GAT layer to map pathway-pathway interactions -- moving towards a foundation model covering all biological pathways that can be broadly applied to many life science tasks.

\bibliographystyle{unsrtnat}

\end{document}